\documentclass[10pt,twocolumn,letterpaper]{article}
\usepackage{cvpr}
\usepackage{times}
\usepackage{epsfig}
\usepackage{graphicx}
\usepackage{amsmath}
\usepackage{amssymb}
\usepackage{bbm}
\usepackage{upgreek}
\usepackage{algorithm}
\usepackage{algorithmic}
\usepackage{authblk}

\usepackage[pagebackref=true,breaklinks=true,letterpaper=true,colorlinks,bookmarks=false]{hyperref}

 \cvprfinalcopy 

\def\cvprPaperID{2256} 

\ifcvprfinal\fi
\begin{document}

\title{ Deep Hierarchical Machine: a Flexible Divide-and-Conquer Architecture}

\author[1]{Shichao Li}
\author[2]{Xin Yang}
\author[1]{Tim Cheng}
\affil[1]{Hong Kong University of Science and Technology}
\affil[2]{Huazhong University of Science and Technology}

\maketitle

\begin{abstract}
   We propose Deep Hierarchical Machine (DHM), a model inspired from the divide-and-conquer strategy while emphasizing representation learning ability and flexibility. A stochastic routing framework as used by recent deep neural decision/regression forests is incorporated, but we remove the need to evaluate unnecessary computation paths by utilizing a different topology and introducing a probabilistic pruning technique. We also show a specified version of DHM (DSHM) for efficiency, which inherits the sparse feature extraction process as in traditional decision tree with pixel-difference feature. To achieve sparse feature extraction, we propose to utilize sparse convolution operation in DSHM and show one possibility of introducing sparse convolution kernels by using local binary convolution layer.  DHM can be applied to both classification and regression problems, and we validate it on standard image classification and face alignment tasks to show its advantages over past architectures.
\end{abstract}

\section{Introduction}

\begin{figure}[t]
	\begin{center}
	\includegraphics[width=1\linewidth]{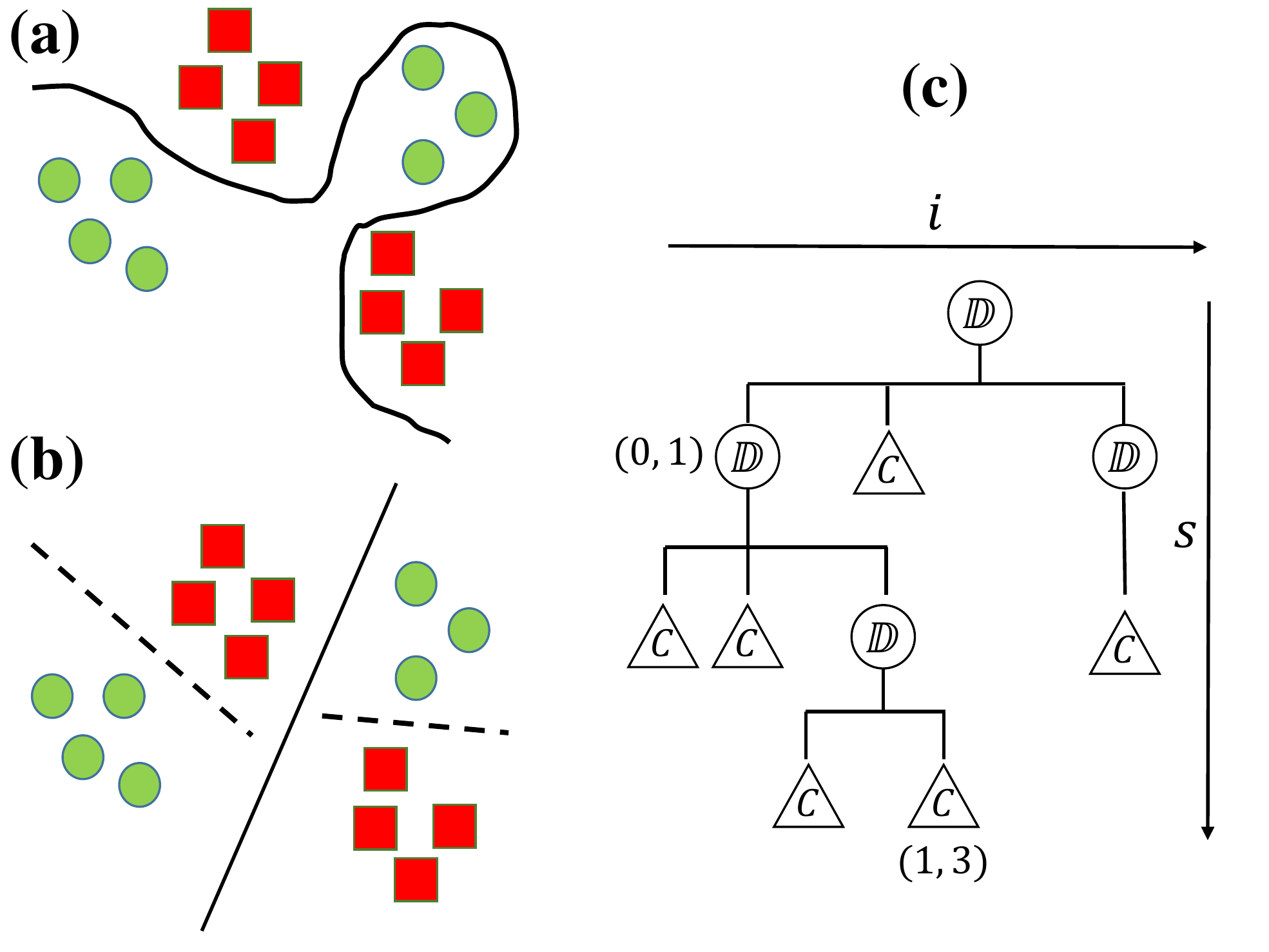}
	\end{center}
	\caption{(a) Finding a decision boundary without dividing the input feature space. (b) Dividing the feature space (solid line) into two sub-regions and find two decision boundaries (dashed line) in each sub-region. (c) Diagram of a Deep Hierarchical Machine. $i$ and $s$ are horizontal index and depth of nodes, respectively.}
	\label{fig1}
\end{figure}

Divide-and-conquer is a widely-adopted problem-solving philosophy which has been demonstrated to be successful in many computer vision tasks, e.g.\ object detection and tracking \cite{5540107} \cite{2015}. Instead of solving a complete and huge problem, divide-and-conquer suggests decomposing the problem into several sub-problems and solving them in different constrained contexts. Figure~\ref{fig1} illustrates this idea with a binary classification problem. Finding a decision boundary in the original problem space is difficult and leads to a sophisticated nonlinear model, but linear decision models could be more easily obtained when solving the sub-problems.

The traditional decision tree, which splits the input feature space at each splitting node and gives the prediction at a leaf node, inherently uses the divide-and-conquer strategy as an inductive bias. The designs of input features and splitting functions are key to the success of this model. Conventional methods usually employ hand-crafted features such as the pixel-difference feature \cite{7130626, 6909637, 6909614, 8031057} and Harr-like feature \cite{321456}. However, the input space for vision tasks are usually high-dimensional and often lead to a huge pool of candidate features and splitting functions that are impractical for an exhaustive evaluation. In practice the huge candidate pool is randomly sampled to form a small candidate set of splitting functions and a local greedy heuristic such as entropy minimization is adopted to choose the "best" splitting function which maximizes data "purity", limiting the representation learning ability of the traditional decision tree.

Deep neural decision forests \cite{7410529} was proposed to enable a decision tree with deep representation learning ability. In \cite{7410529}, the outputs of the last fully connected layer of a CNN are utilized as stochastic splitting functions. A global loss function is differentiable with respect to the network parameters in this framework, enabling greater representation learning ability than the local greedy heuristics in conventional decision trees. Deep regression forests \cite{123456} was later proposed for regression problems based on the general framework of \cite{7410529}. However, the success in introducing representation learning ability comes with the price of transforming decision trees into stochastic trees which make “soft” decision at each splitting node. As a result, all splitting functions have to be evaluated as every leaf node contributes to the final prediction, yielding a significant time cost. Pruning branches that contribute little to the final prediction should effectively reduce the computational cost with little accuracy degradation. Unfortunately, the network topology used in previous works \cite{7410529, 123456} requires a complete forward pass of the entire CNN to compute the routing probability for each splitting node, making pruning impractical.

A major advantage of the divide-and-conquer strategy (e.g.\ random forests) is its high efficiency in many time-constraint vision tasks such as face detection and face alignment. Simple and ultrafast-to-compute features such as pixel difference, only extract sparse information (e.g. two pixels) from the image space. However, existing deep neural decision/regression forests \cite{7410529, 123456}  completely ignore the computational complexity of splitting nodes and in turn greatly limit their efficiency. 

In this work, we propose a general tree-like model architecture, named Deep Hierarchical Machine (DHM), which utilizes a flexible model topology to decouple the evaluation of splitting nodes and a probabilistic pruning strategy to avoid the evaluation of unnecessary paths. For the splitting nodes, we also explore the feasibility of inheriting the sparse feature extraction process (i.e.\ the pixel-difference feature) of the traditional random forests and design a deep sparse hierarchical machine (DSHM) for high efficiency. We evaluate our method on standard image classification and facial landmark coordinate regression tasks and show its effectiveness. Our implementation can be easily incorporated into any deep learning frameworks and the source code and pre-trained models will be available on the website\footnote{The website address is currently unavailable.}. In summary, our contributions are:

\begin{itemize}
	\item[1.] We propose Deep Hierarchical Machine (DHM) with a flexible model topology and probabilistic pruning strategy to avoid evaluating unnecessary paths. The DHM enjoys a unified framework for both classification and regression tasks.
	\item[2.] We introduce sparse feature extraction process into DHM, which to our best knowledge is the first attempt to mimic traditional decision trees with pixel-difference feature in deep models.
	\item[3.] For the first time, we study using deep regression tree for a multi-task problem, i.e., regressing multiple facial landmarks. 
\end{itemize}


\section{Related works}

We list three related topics in this section to show two trends in the computer vison community. The first is the migration from hand-crafted features towards deep representation learning for divide-and-conquer models, the second is the generalization of sparse pixel-based features to sparse operation in deep convolutional neural networks.

\subsection{Traditional divide-and-conquer models}

Traditional decision trees or random forests \cite{97458, 2013978} can be naturally viewed as divide-and-conquer models, where each non-leaf node in the tree splits the input feature space and route the input deterministically to one of its children nodes. These models employ a greedy heuristic training procedure which randomly samples a huge pool of candidate splitting functions to minimize a local loss function. The parameter sampling procedure is sub-optimal compared to using optimization techniques, which in combination of the hand-crafted nature of the used features, limit these models' representation learning ability. Hierarchal mixture of experts \cite{181214} also partitions the problem space in a tree-like structure using some gating models and distribute inputs to each expert model with a probability. A global maximum likelihood estimation task was formulated under a generative model framework, and EM algorithm was proposed to optimize linear gating and expert models. This work inspires our methodology but deep representation learning and probabilistic pruning was not studied at that time.

\subsection{Deep decision/regression tree}

\cite{7410529, 123456} proposed to extract deep features to divide the problem space and use simple probabilistic distribution at leaf nodes. These models enabled traditional decision/regression trees with deep representation learning ability. Leaf node update rules were proposed based on convex optimization techniques, and they out-performed deep models without divide-and-conquer strategy. However, since the last layer of a deep model was used to divide the problem space, every path in the tree needs to be computed. Even when a branch of computation contributes little to the final prediction, it stills need evaluation because each splitting node requires the full forward-pass of the deep neural network. A model structure where each splitting node is separately evaluated was used \cite{7780963} for depth estimation, but a general framework was missing and the effect of computation path pruning was not investigated.

\subsection{Sparse feature extraction}

Pixel-difference feature is a special type of hand-crafted feature where only several pixels from an input are considered during its evaluation. They are thus efficient to compute and succeeded in computer vision tasks such as face detection \cite{7130626}, face alignment \cite{6909614, 6909637, 6248015, 8031057, 109122}, pose estimation \cite{5995316, 159232} and body part classification \cite{7298672}. These features were also naturally incorporated into decision/regression trees to divide the input feature space. A counterpart of sparse feature extraction process in CNNs is sparse convolution where the few non-zero entries in the convolution kernel determine the feature extraction process. To obtain a sparse convolution kernel, sparse decomposition \cite{7298681} and pruning \cite{111222} techniques were proposed to sparsify a pre-trained dense CNN. \cite{8099939} proposed an alternative where random sparse kernel was initialized before the training process. While they focus on speeding up CNNs, there have not been study on using these sparse convolutional layers in problem space dividing process, as traditional pixel-difference feature was used in decision trees.


\section{Methodology}

We first formulate the general DHM based on a hierarchical mixture of experts (HME) framework, then we specify the model for classification and regression experiments.

\subsection{General framework of DHM}
The general divide-and-conquer strategy consists of multiple levels of dividing operations and one final conquering step. The computation process is depicted as a tree where all leaf nodes are called conquering nodes while the others are named as dividing nodes. We index a node by a tuple subscript $(i,s)$ where $s$ denotes the vertical stage depth (see Figure~\ref{fig1}) and $i$ denotes the horizontal index of the node. Every node has a non-negative integer number of children nodes, which forms a sequence $\mathcal{K}_{i,s} = \{\mathcal{K}_{i,s}^{1}, \mathcal{K}_{i,s}^{2}, ..., \mathcal{K}_{i,s}^{|\mathcal{K}_{i,s}|}\} $. Each node has exactly one input $\mathcal{I}_{i,s} $ and one output $\mathcal{O}_{i,s} $. 

A dividing node $\mathcal{D}_{i,s}$ is composed of a tuple of functions $(\mathcal{R}_{i,s}, \mathcal{M}_{i,s} )$. The first function is called the recommendation function which judges the node input and gives the recommendation score vector $\mathbf{s}_{i,s} = \mathcal{R}_{i,s}(\mathcal{I}_{i,s})$ whose length equals the children sequence length $|\mathcal{K}_{i,s} |$ and the $j$th entry $\mathbf{s}_{i,s}(j)$ is a real number associated with the $j$th child node. We require
\begin{equation}
\label{eq:requirement}
0\leq\mathbf{s}_{i,s}(j)\leq1, \sum_{j=1}^{|\mathcal{K}_{i,s}|}\mathbf{s}_{i,s}(j)=1
\end{equation}
so that $\mathbf{s}_{i,s}(j)$ can be considered as the significance or probability of recommending the input $\mathcal{I}_{i,s} $ to the $j$th child node. The second function $\mathcal{M}_{i,s}$ is called mapping function and maps the input to form the output of the node $\mathcal{O}_{i,s} = \mathcal{M}_{i,s}(\mathcal{I}_{i,s})$, which is allowed to be copied and sent to all its children nodes $\mathcal{K}_{i,s}$. 

We name the unique path from the root node to one conquering (leaf) node a computation path $\mathcal{P}_{i,s}$. Each conquering node only stores one function $\mathcal{M}_{i,s}$ that maps its input into a prediction vector $\mathbf{p}_{i,s} =  \mathcal{M}_{i,s}(\mathcal{I}_{i,s})$, which is considered the termination of its computation path. To get the final prediction $ \mathbf{P}$, each conquering node contributes its output weighted by the probability of taking its computation path as  
\begin{equation}
	\mathbf{P} = \sum_{(i,s)\in\mathcal{N}_{c}}w_{i,s}\mathbf{p}_{i,s}
\end{equation}
and $\mathcal{N}_{c}$ is the set of all conquering nodes.
The weight can be obtained by multiplying all the recommendation scores along the path given by each dividing node. Assume the path $\mathcal{P}_{i,s}$ consists of a sequence of $s$ dividing nodes and one conquering node as $\{\mathcal{D}_{i_1,s_1}^{j_1}, \mathcal{D}_{i_2,s_2}^{j_2}, \ldots, {\mathcal{C}_{i,s}}\}$, where the superscript for a dividing node denotes which child node to choose. Then the weight can be expressed as
\begin{equation}
w_{i,s} = \prod_{m=1}^{s}\mathbf{s}_{i_m,s_m}(j_m)
\end{equation} 
Note that the weights of all conquering nodes sum to 1 due to (\ref{eq:requirement}) and the final prediction is hence a convex combination of all the outputs of conquering nodes. In addition, we assume every function mentioned above is a differentiable function parametrized by $\boldsymbol{\uptheta}_{i,s}^{\mathcal{R}}$ or
$\boldsymbol{\uptheta}_{i,s}^{\mathcal{M}}$ for recommendation or mapping function at node $(i,s)$. Thus the final prediction is a differentiable function with respect to all the parameters which we omit above to ensure clarity. A loss function defined upon the final prediction can hence be optimized with back-propagation algorithm and benefit from some frameworks that provide automatic differentiation.
\begin{figure}[h]
	\begin{center}
		\includegraphics[width=1\linewidth]{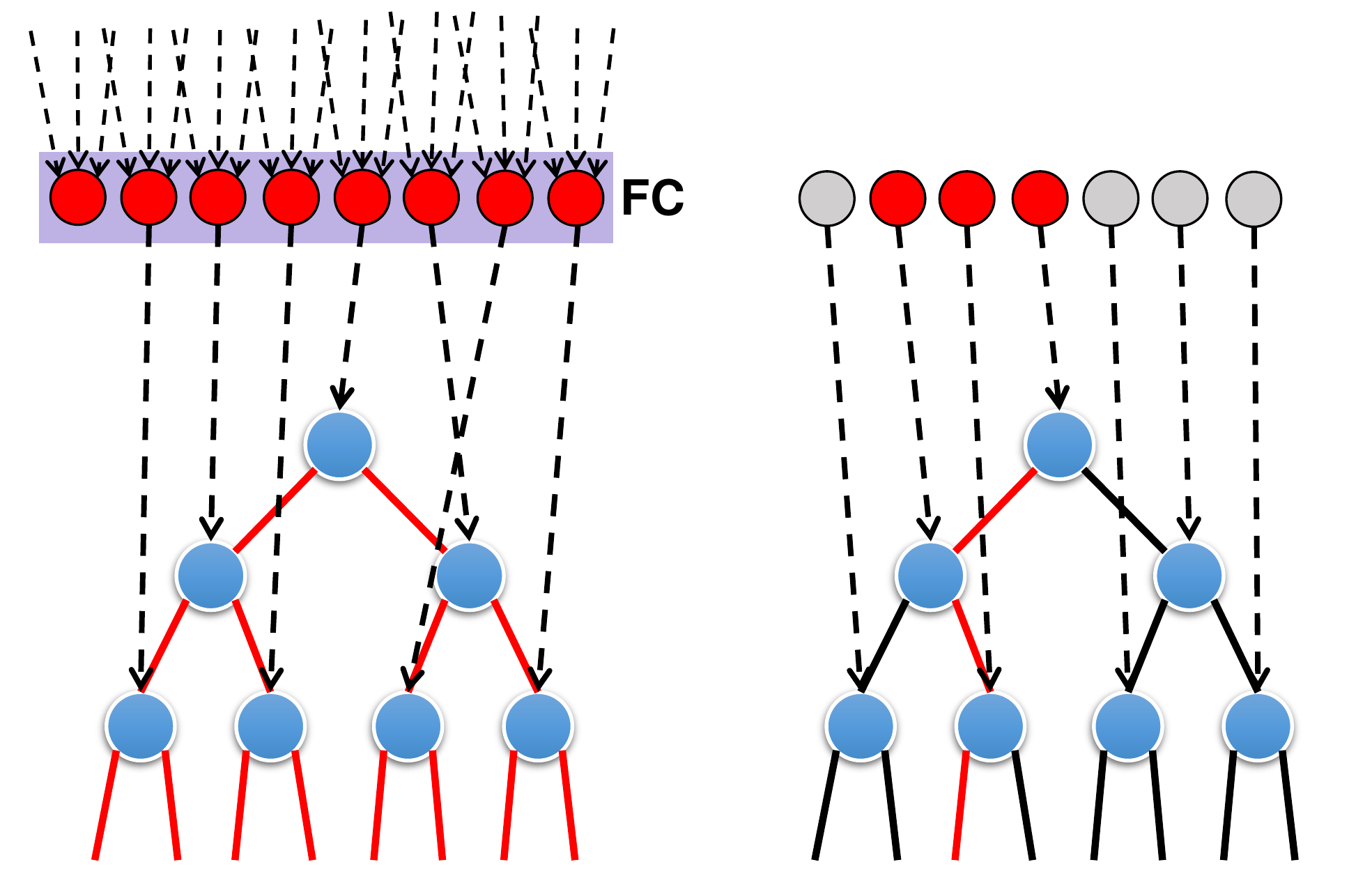}
	\end{center}
	\caption{Past (left) and our (right) methods of setting recommendation functions. Blue circles are dividing nodes and dashed lines define the recommendation functions from coupled or separated models. Red circles and edges are evaluated models and chosen paths at run time while gray circles and black edges are not evaluated or chosen.}
	\label{topology}
\end{figure}

A flexible feature in this framework is that, the recommendation functions $\mathcal{R}_{i,s}$ are in general not coupled with each other. \cite{7410529, 123456} pass the last fully-connected layer to sigmoid gates, whose results are used as recommendation scores in the dividing nodes (Figure~\ref{topology} left). In this way all recommendation functions are evaluated simultaneously to give probabilities of taking all computation paths, even when most of the paths contribute little to the final results. On the other hand, our framework allows separation of the recommendation functions (Figure~\ref{topology} right) so that we can avoid evaluating unnecessary computation paths.

We define a Probabilistic Pruning (PP) strategy based on the separability of the recommendation functions.
Starting from the root dividing node, its children node will not be visited if their corresponding recommendation scores are lower than a pruning threshold $\mathcal{P}_{th} $. This process recursively applies to its descendant dividing nodes and finally the more important computation paths are preserved. The process is depicted Algorithm~\ref{PP}.

\begin{algorithm}
	\caption{Probabilistic Pruning for $\mathcal{D}_{i,s}$}
	\begin{algorithmic} [1]
		\REQUIRE $i,s, \mathcal{P}_{th} \geq 0, |\mathcal{K}_{i,s}| \geq 1$
		\STATE $sum = 0$
      \FOR{$1 \leq j \leq |\mathcal{K}_{i,s}|$}
      \IF{$\mathbf{s}_{i,s}(j) < \mathcal{P}_{th} $} \STATE $\mathbf{s}_{i,s}(j) \leftarrow 0$ 
      \ELSE
      \STATE $sum = sum + \mathbf{s}_{i,s}(j)$
      \ENDIF
      \ENDFOR	
	  \STATE $\mathbf{s}_{i,s} = \mathbf{s}_{i,s}/sum $
      \FOR{$1 \leq j \leq |\mathcal{K}_{i,s}|$}
      \IF{$\mathbf{s}_{i,s}(j) > 0 $}
      \STATE Do Probabilistic Pruning for $\mathcal{K}_{i,s}^{j}$  
      \ENDIF
      \ENDFOR	  
	\end{algorithmic}
	\label{PP}
\end{algorithm}

Up to now, this model with decoupled recommendation functions is called Deep Hierarchical Machine (DHM) and Deep Sparse Hierarchical Machine (DSHM) is a specific form of DHM where a sparse feature extractor is used inside a node. For instance, $\mathcal{R}_{i,s}(\mathcal{I}_{i,s}) = \mathcal{R}_{i,s}(\mathcal{G}(\mathcal{I}_{i,s}))$ where $\mathcal{G}$ only considers a small portion of input $\mathcal{I}_{i,s}$.

\subsection{Classification}
For classification problem, the output $\mathbf{p}_{i,s}$ for each conquering node $\mathcal{C}_{i,s}$ is a discrete probability distribution vector whose length equals the number of classes. The $y$th entry $\mathbf{p}_{i,s}(y)$ gives the probability $\mathbb{P}(y|\mathcal{I}_{0,0}) $ that the root node input $\mathcal{I}_{0,0}$ belongs to class $y$ .

To train the model, we adopt the probabilistic generative model formulation \cite{181214} which leads to a maximum likelihood optimization problem. For one training instance which is composed of an input vector and a class label $\{\mathbf{x}_i, y_i\}$, the likelihood of generating it is, 
\begin{equation}
\mathbb{P}(y_i|\mathbf{x}_i) = \sum_{(i,s)\in\mathcal{N}_{c}} \prod_{m=1}^{s}\mathbf{s}_{i_m,s_m}(j_m)\mathbf{p}_{i,s}(y_i)
\end{equation}
The optimization target is to minimize the negative log-likelihood loss over the whole training set containing $N$ instances $\mathbb{D} = \{\mathbf{x}_i,y_i\}_{i=1}^{N} $,
\begin{equation}
L(\mathbb{D}) = -\sum_{i=1}^{N}\log(\mathbb{P}(y_i|\mathbf{x}_i))
\end{equation}

In this study, we constrain each dividing node to have exactly two children since we do not assume any prior knowledge on how many parts the input feature space should to be split into. We also assume a full binary-tree structure for simplicity. If some application-specific information such as clustering results are available, the tree structure can be adjusted accordingly. In the case of full binary tree, we can index each node with a single non-negative integer $i$ for convenience. The recommendation function in each diving-node only needs to give a 2-vector $\mathbf{s}_{i}$ and we use the short-hand $ s_{i}$ to denote the probability the current dividing node input $ \mathcal{I}_{i}$ is recommended to the left sub-tree. For a dividing node $\mathcal{D}_i$, we denote nodes in its left and right sub-trees as node sets ${\mathcal{D}_i^l}$ and ${\mathcal{D}_i^r}$, respectively. Then the probability of recommending the input $\mathbf{x}$ to a conquering node $\mathcal{C}_{i}$ can be expressed as,
\begin{equation}
\mathbb{P}(\mathcal{C}_{i}|\mathbf{x}) = \prod_{\mathcal{D}_{j}\in\mathcal{N}_{d}}
s_j^{\mathbbm{1}(\mathcal{C}_{i}\in{\mathcal{D}_j^l})} {(1 - s_j)}^{\mathbbm{1}(\mathcal{C}_{i}\in{
		\mathcal{D}_j^r})}
\end{equation}
where $\mathcal{N}_{d}$ is the set of all dividing nodes and $\mathbbm{1} $ is an indicator variable for the expression inside the parenthesis to hold. For the classification experiments we use the simplest conquering strategy for each conquering node as in \cite{7410529}, where each conquering node gives a constant probability distribution $\mathbf{p}_{i}$.  
The loss function is differentiable with respect to each $s_i$, and the gradient for this full binary tree structure $\frac{\partial L(\mathbb{D})}{\partial{s_i}}$ is \cite{7410529, 123456, 7780963},

\begin{equation}
\sum_{t = 1}^{N}
 (\frac{{\sum_{\mathcal{C}_j\in \mathcal{D}_i^l}\mathbf{p}_{j}(y_{t})\mathbb{P}(\mathcal{C}_{j}|\mathbf{x}_t)} }{s_{i}\mathbb{P}(y_t|\mathbf{x}_t)}
 - \frac{\sum_{\mathcal{C}_j\in \mathcal{D}_i^r}\mathbf{p}_{j}(y_{t})\mathbb{P}(\mathcal{C}_{j}|\mathbf{x}_t)}{(1-s_{i})\mathbb{P}(y_t|\mathbf{x}_t)})
\end{equation}
\begin{figure}[t]
	\begin{center}
		\includegraphics[width=1\linewidth]{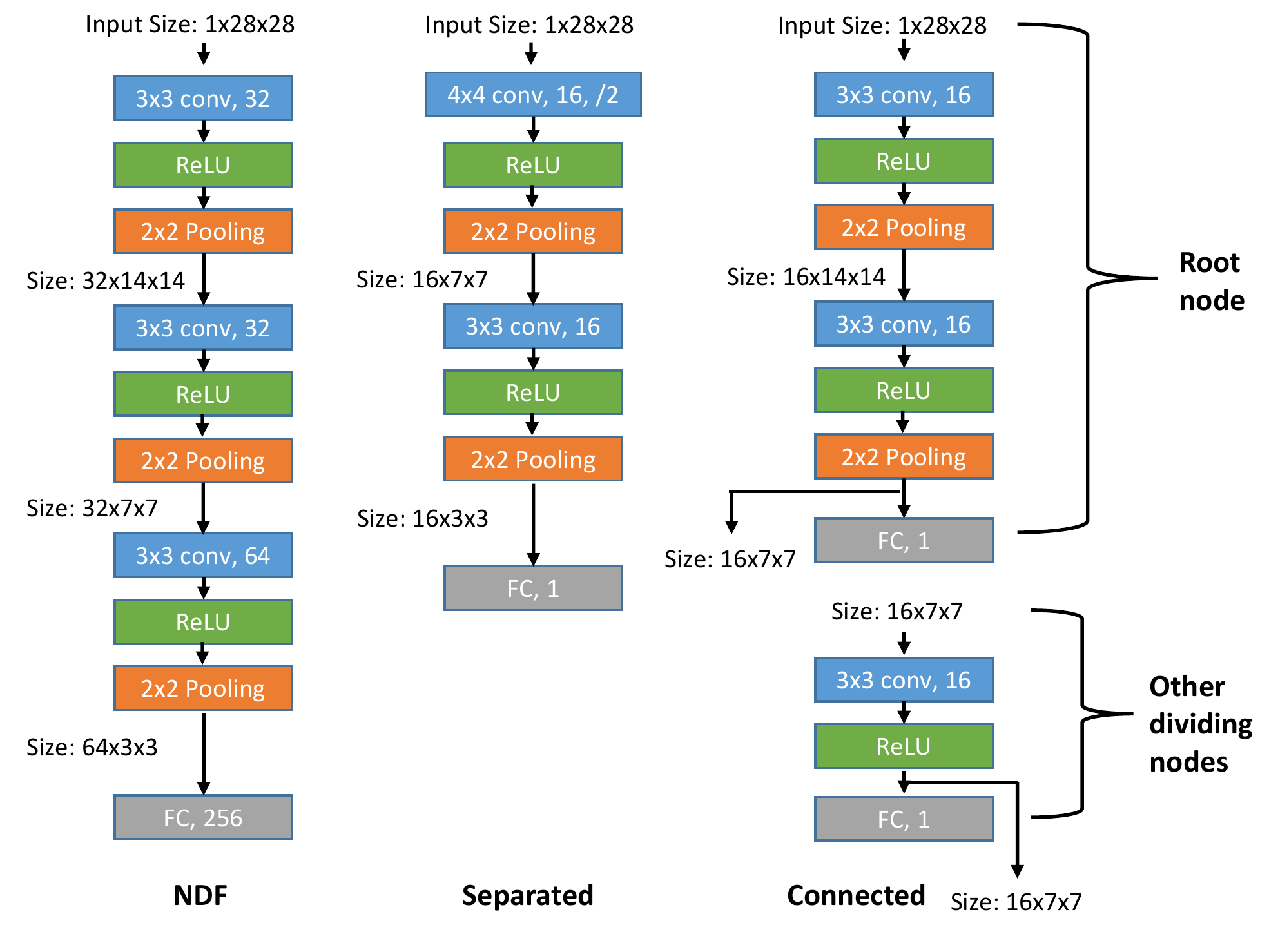}
	\end{center}
	\caption{Different architectures used in the experiments. Left, original neural decision forest architecture. Middle, the architecture of a dividing node in DHM with separated recommendation functions. Right, the architecture of a root node and dividing nodes in DHM with connected recommendation functions.}
	\label{model}
\end{figure}
This gradient can be passed backward into each dividing node to train its function parameters. Note that in our framework each $\mathcal{D}_i$ is generally decoupled with each other while in \cite{7410529} and \cite{123456} all $\mathcal{D}_i$ come from the last layer of a deep model and are hence coupled. When the dividing nodes are fixed, the distribution at each conquering node can be updated iteratively \cite{7410529},
\begin{equation}
\mathbf{p}_j^{t+1}(y) = \frac{1}{Q_{j}^{t}}
\sum_{i=0}^{N}
\frac{\mathbbm{1}(y_{i} = y)
	\mathbf{p}_{j}^{t}(y_{i})\mathbb{P}(\mathcal{C}_{j}|\mathbf{x}_i)
	}{\mathbb{P}(y_i|\mathbf{x}_i)}
\label{update}
\end{equation}
where  $Q_{j}^{t}$ is a normalization factor to ensure $\sum_{y=1}^{|\mathbf{p}_j|}\mathbf{p}_j^{t+1}(y) = 1$. The backward propagation and the conquering nodes update are carried out alternately to train the model. 
\subsection{Regression}
For regression problems, the output of a conquering node $\mathcal{C}_{i,s}$ is also a real-valued vector $\mathbf{p}_{i,s}$ but the entries do not necessarily sum to 1. The final prediction vector $\mathbf{P}_{i}$ for input $\mathbf{x}_i$ is,
\begin{equation}
\mathbf{P}_{i} = \sum_{(i,s)\in\mathcal{N}_{c}} \prod_{m=1}^{s}\mathbf{s}_{i_m,s_m}(j_m)\mathbf{p}_{i,s}
\end{equation}
For a multi-task regression dataset with $N$ instances $\mathbb{D} = \{\mathbf{x}_i,\mathbf{y}_i\}_{i=1}^{N} $, we directly use the squared loss function, 
\begin{equation}
	L(\mathbb{D}) =\frac{1}{2} \sum_{i=1}^{N}\lvert\lvert\mathbf{P}_{i}-\mathbf{y}_{i}\rvert\rvert^2
\end{equation}
which was also used in the mixture of experts framework \cite{0269282}. Here we use the same full binary tree structure and assume simple conquering nodes which have constant mapping functions just as the classification case. Similarly, $\frac{\partial L(\mathbb{D})}{\partial{s_i}}$ is computed as,
\begin{equation}
\frac{\partial L(\mathbb{D})}{\partial{s_i}} = \sum_{t = 1}^{N}(\mathbf{P}_{t} - \mathbf{y}_{t})^T
(\frac{\mathbf{A}_l}{s_{i}}
- \frac{\mathbf{A}_r}{(1-s_{i})})
\end{equation}
where $\mathbf{A}_l = \sum_{\mathcal{C}_j\in \mathcal{D}_i^l}\mathbb{P}(\mathcal{C}_{j}|\mathbf{x}_i)\mathbf{p}_{j} $ and $\mathbf{A}_r = \sum_{\mathcal{C}_j\in \mathcal{D}_i^r}\mathbb{P}(\mathcal{C}_{j}|\mathbf{x}_i)\mathbf{p}_{j} $.
Similar to \ref{update}, we update the conquering node prediction as
\begin{equation}
\mathbf{p}_j^{t+1} = 
\frac{\sum_{i=0}^{N}\mathbf{y}_i\mathbb{P}(\mathcal{C}_{j}|\mathbf{x}_i)
}{\sum_{i=0}^{N}\mathbb{P}(\mathcal{C}_{j}|\mathbf{x}_i)}
\label{update_reg}
\end{equation}
This update rule is inspired from traditional regression trees which compute an average of target vectors that are routed to a leaf node. Here the target vectors are weighted by how likely it is recommended into this conquering node.
\section{Experiments}
\subsection{Classification for MNIST}
We start with an illustration using MNIST. We compare the model architecture of \cite{7410529, 123456} with two variants of our proposed DHM as shown in Figure~\ref{model}. The original architecture \cite{7410529, 123456} is denoted as NDF. NDF passes some randomly chosen outputs from the last fully-connected layer to sigmoid gates, whose outputs are used as the recommendation scores $s_{i}$ of each dividing node. The other two structures are detailed in the following subsections.

The MNIST data set contains 60000 training images and 10000 testing images of size 28 by 28 \footnote{\url{https://pytorch.org/docs/0.4.0/_modules/torchvision/datasets/mnist.html}}. During the experiment, binary tree depth and tree number are set to 7 and 1, respectively. Adam optimizer is used with learning rate specified as 0.001. Batch size is set to 500 and the training time is fixed to 50 epochs. Every experiment is repeated 10 times and averaged results with standard deviation are reported.

\subsubsection{Separated Recommendation Functions}

This type of DHM separates each dividing node's input and output, as shown in the middle column of Figure~\ref{model}. Each dividing node processes the raw input image and produces a single number after the fully-connected layer, which is passed through a sigmoid function to give $s_i$. One can think of this structure as the mapping functions for all dividing nodes are identity mappings $\mathcal{M}_{i,s}(\mathcal{I}_{i,s}) = \mathcal{I}_{i,s} $. We denote this type as DHM (separated). The final test accuracy of this and other types of models are summarized in Table~\ref{pruning_ac}. In addition, we estimate the computation load by the number of multiplication (NOM) operation needed in the convolution and linear layers, which is shown in Table~\ref{pruning_sp}.
\begin{table}
	\begin{center}
		\begin{tabular}{l|c|c}
			\hline
			Method & Accuracy  & After Pruning\\
			\hline\hline
			NDF & 0.9896$\pm$0.0023 & Not Able \\
			\hline 
			DHM (separated) & 0.9860$\pm$0.0010 & 0.9853$\pm$0.0010\\
			DHM (connected) & 0.9861$\pm$0.0019 &  0.9856$\pm$0.0020\\
			\hline
		\end{tabular}
		
	\end{center}
	\caption{The test accuracy of different models before and after probabilistic pruning.}
	\label{pruning_ac}
\end{table}

\begin{figure*}[t]
	\begin{center}
		\includegraphics[width=1\linewidth]{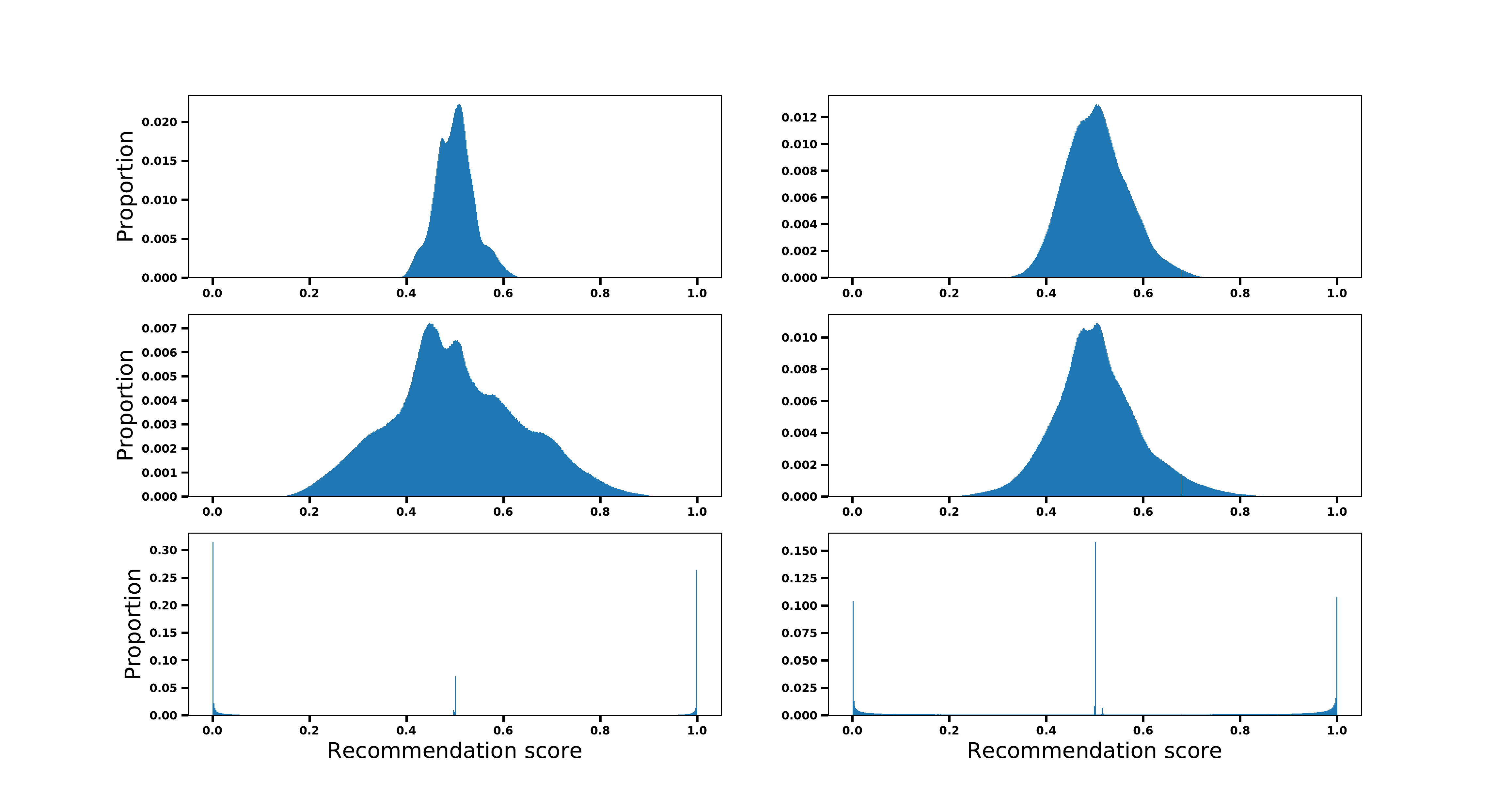}
	\end{center}
	\caption{Bar plots showing the evolution of the distribution of recommendation scores during training. The left column depicts results from NDF while the right shows that of a DHM with separated recommendation functions. The first to the third rows are taken after 1, 7, and 500 iterations of training.}
	\label{bar}
\end{figure*}

\subsubsection{Deeper Feature Along the Path}
In this type of architecture, the root dividing node does more initial processing and reduces the size of the input images (See the right column of Figure~\ref{model}). Other dividing nodes pass the processed feature maps to its children dividing nodes as inputs. Every dividing node also sends their flattened outputs to a linear and sigmoid layer to produce $s_i$. The mapping function in this case can be seen as the local network without the last fully-connected layer. The intuition to use this topology is that the node input at larger depth will pass more dividing nodes and be processed more times. This type of model is denoted as DHM (connected). 

\begin{table}
	\begin{center}
		\begin{tabular}{l|c|c}
			\hline
			Method & NOM  & After Pruning\\
			\hline\hline
			NDF & 3.08M & Not Able \\
			\hline 
			DHM (separated) & 20.9M & 1.14M\\
			DHM (connected) & 15.1M & 1.25M\\
			\hline
		\end{tabular}
		
	\end{center}
	\caption{The number of multiplication (NOM) before and after probabilistic pruning.}
	\label{pruning_sp}
\end{table}

\subsubsection{Probabilistic Pruning}
The distribution of $s_i$ during the training process is shown in Figure~\ref{bar}. Every bar plot contains 500 bins to quantize all dividing nodes' $s_i$ values from 60000 training images. After initialization the distribution is centered around 0.5 while after longer training time, the dividing nodes are more decisive to recommend their inputs.  When $s_i$ is very close to 1 or 0, the contribution from one of the two sub-trees is too low to be worthwhile for extra evaluation. This motivates the Probabilistic Pruning (PP) strategy which gives up evaluation of a sub-tree dynamically if the recommendation score of entering it is too low. NDF does not support PP even if the distribution strongly encourages it (see Figure~\ref{bar} left), since all dividing nodes are coupled to the last fully-connected layer of the network. On the other hand, DHM can support PP naturally. In the experiment, we set the pruning threshold as 0.5 so that only one computation path is taken for every input image. The resulting test accuracy and NOM are shown in Table~\ref{pruning_ac} and Table~\ref{pruning_sp}, respectively. Applying PP only sacrifices the testing accuracy negligibly but the computational cost is reduced from exponential to linear since now the most significant computation path determines the result. These results prove that DHM can take advantage of the distribution of recommendation scores.

\begin{figure}
	\begin{center}
		\includegraphics[width=1\linewidth]{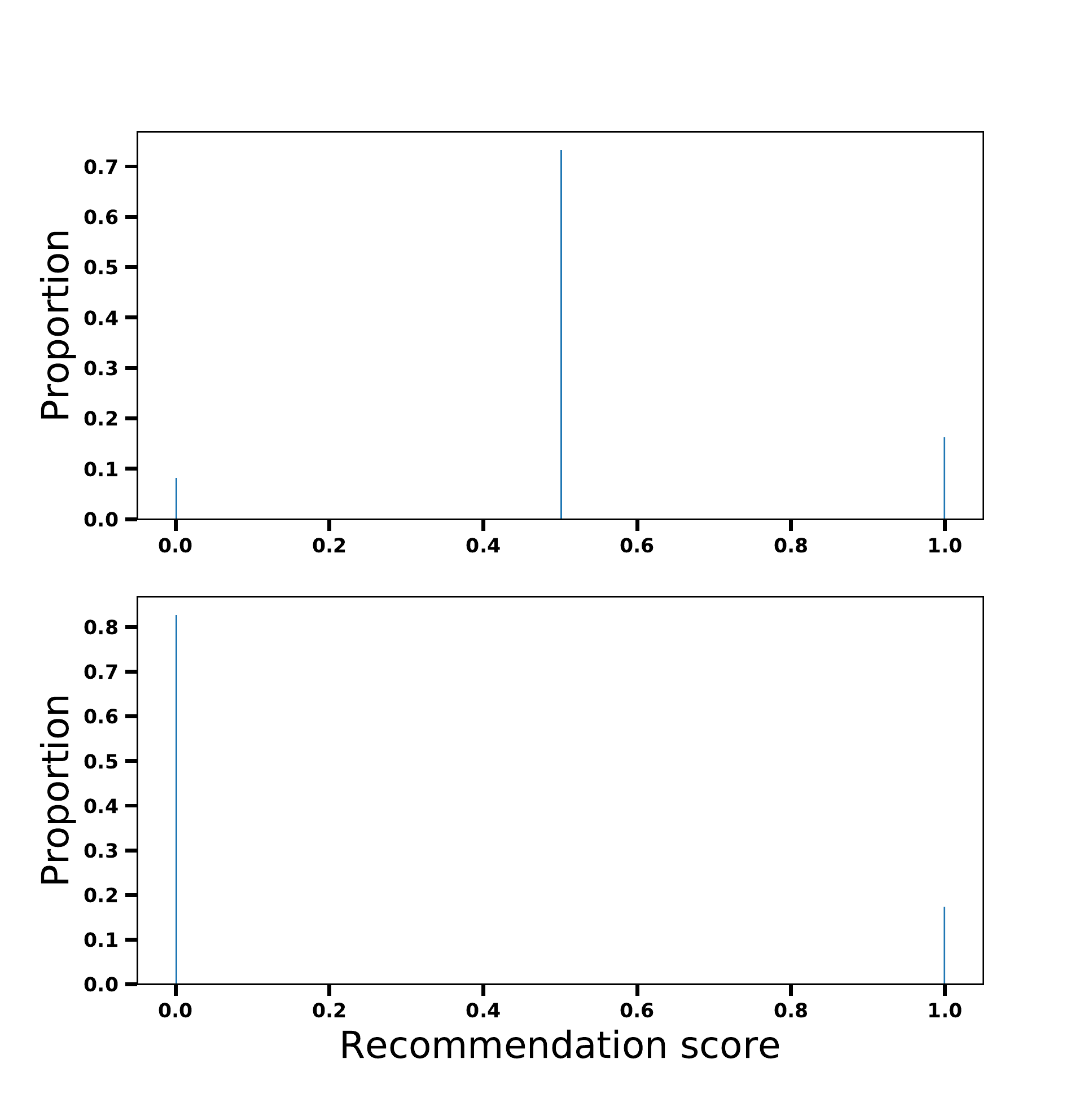}
	\end{center}
	\caption{The distribution of recommendations scores over the test set before (top) and after (bottom) probabilistic pruning.}
	\label{pruning}
\end{figure}

The recommendation scores distribution for testing images before and after pruning is shown in Figure~\ref{pruning}. Surprisingly, when a large amount of "hesitating" dividing nodes are deterministically given which child-node to use, the accuracy was not affected significantly. 

\subsubsection{Adding Sparsity}
Here we use local binary convolution \cite{8099939} to add sparse feature extraction process into DHM, making it DSHM. Every original convolution layer is replaced by two convolution layers and a ReLU gate. The first convolution layer is fixed and does not introduce any learnable parameters. The output feature maps of the first layer is passed to the ReLU gate, whose outputs are linear combined by the second 1 by 1 convolution layer. During initialization, some entries in the convolution kernel of the first layer are randomly assigned to be zero. The remaining entries are randomly assigned to 1 or -1 with probability 0.5 for each option. The percentage of non-zero entries in the fixed convolution kernel is defined as the sparsity level. In the experiment, we use 16 intermediate channels (output feature map number of the first layer) for all local binary convolution layers. DHM (separated) is used and other network parameters are consistent with the former experiments without sparse convolution layer.

\begin{table}
	\begin{center}
		\begin{tabular}{l|c|c}
			\hline
			Sparcity level & NOM  & Accuracy\\
			\hline\hline
			0.3 & 8.05M & 0.9719$\pm$0.0011\\
			0.5 & 8.05M & 0.9725$\pm$0.0021\\
			0.7 & 8.05M & 0.9709$\pm$0.0024\\
			\hline
		\end{tabular}
	\end{center}
	\caption{The number of multiplication (NOM) and test accuracy of DSHM with different sparsity levels. Note here the NOM does not consider PP and should be compared with DHM without PP.}
	\label{sparse}
\end{table}

The resulting test accuracy and NOM is shown in Table \ref{sparse}. Since convolution with binary kernel can be implemented by addition and subtraction, the required NOM is further reduced. This experiment shows sparse feature extraction process can be seamlessly incorporated into DHM, which can be used in devices with limited computational resources.  

\subsection{Cascaded regression with DHM}
\begin{figure}[h]
	\begin{center}
		\includegraphics[width=1\linewidth]{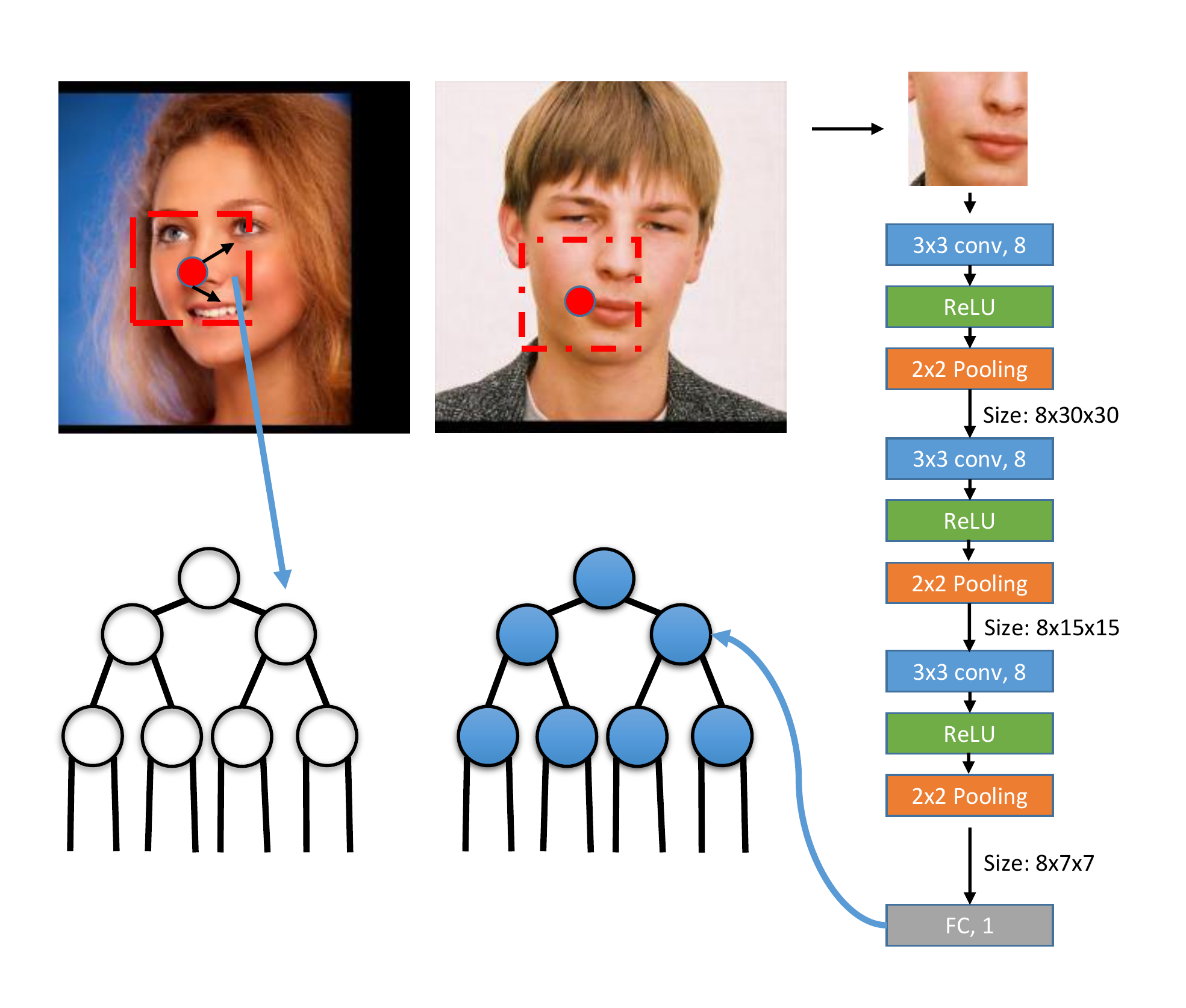}
	\end{center}
	\caption{Feature extraction in a traditional regression tree based on pixel difference feature (left) and our DHM (right). }
	\label{face}
\end{figure}

Here we compare DHM with NDF architecture for a regression task, i.e., cascaded regression based face alignment. For an input image $\mathbf{x}_i$, the goal of face alignment is to predict the facial landmark position vector $\mathbf{y}_i$. Cascaded regression method starts with an initialized shape $\hat{\mathbf{y}}_0$ and use a cascade of regressors to update the estimated facial shape stage by stage. The final prediction $\hat{\mathbf{y}} = \hat{\mathbf{y}}_0 + \sum_{t=1}^{K}\Delta \mathbf{y}_{t} $ where $K$ is the total stage number and $\Delta\mathbf{y}_{t}$ is the shape update at stage $ t$. In \cite{6909637, 8031057}, every regressor was an ensemble of regression trees  whose leaf nodes give the shape update vector. Every splitting node in the regression tree locates two pixels around one current estimated landmark, whose difference was used to route the input in a hard-splitting manner. (see Figure~\ref{face}) 

\begin{figure}
	\begin{center}
		\includegraphics[width=0.8\linewidth]{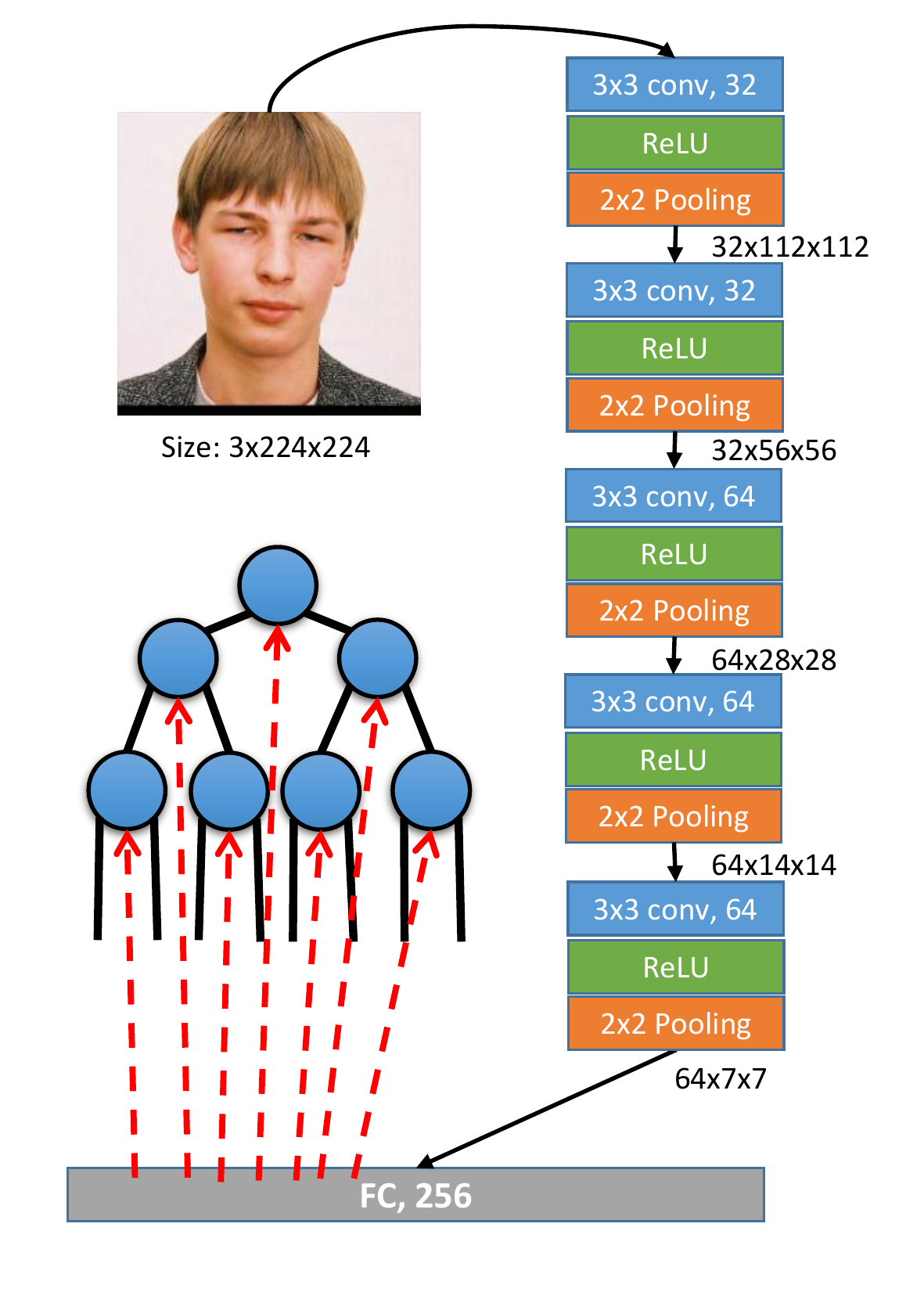}
	\end{center}
	\caption{Face alignment experiment with traditional NDF architecture. }
	\label{faceglobal}
\end{figure}

We replace the the traditional regression trees with our DHMs that use a full binary tree structure so as to extend \cite{6909637} with deep representation learning ability. During initialization, every dividing node is randomly assigned a landmark index. The input to a dividing node is then a cropped region centered around its indexed landmark. In the experiment we use a crop size of 60 by 60 and a simple CNN to compute the recommendation score, whose structure is shown in Figure~\ref{face}. The comparison group uses traditional NDF architecture and we feed entire image as input (see Figure~\ref{faceglobal}). Similarly, every conquering node store a shape update vector as in \cite{6909637, 8031057} and (\ref{update_reg}) is used to update them.

\begin{table}
	\begin{center}
		\begin{tabular}{l|c|c}
			\hline
			Method & NOM (After Pruning) & Error (After Pruning) \\
			\hline\hline
			NDF & 254M (Not able) & 0.0643 (Not able)\\
			DHM & 228M (35.6M) & 0.0628 (0.06382)\\
			\hline
		\end{tabular}
	\end{center}
	\caption{The comparison of traditional NDF architecture and our DHM for regression task. Numbers in the parentheses give the results with PP and only one computational path was taken.}
	\label{facetable}
\end{table}

\begin{figure}
	\begin{center}
		\includegraphics[width=0.9\linewidth]{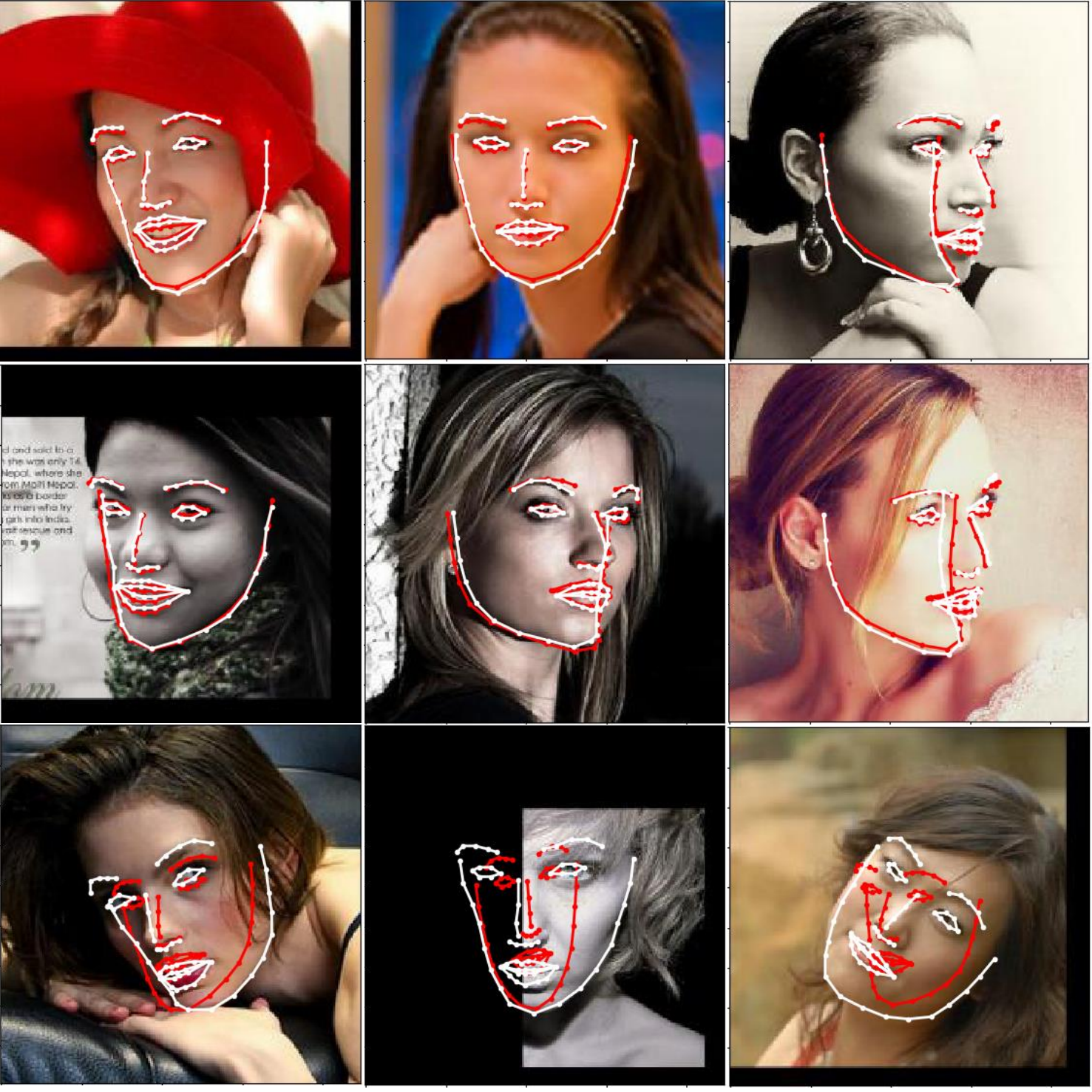}
	\end{center}
	\caption{Face alignment testing results with DHM. White and red dots are ground truth and prediction, respectively.}
	\label{result}
\end{figure}

We use a large scale synthetic 3D face alignment dataset 300W-LP \cite{146155} for training and the AFLW3D dataset (re-annotated by \cite{8237378}) for testing. We use 57559 training images in 300W-LP and the whole 1998 images in the AFLW3D for testing. The images are cropped and resized to 224 by 224 patch using the same initial processing procedure in \cite{8237378}. To rule out the influences of face detectors as mentioned in \cite{7384759}, a bounding box for a face is assumed to be centered at the centroid of the facial landmarks and encloses all the facial landmarks inside. We use the same error metric as \cite{8237378} where the landmark prediction error is normalized by the bounding box size. In the experiment we use a cascade length of 10 and tree depth of 5 and in each stage we use an ensemble of 5 DHMs. We use the ADAM optimizer with a learning rate at 0.01 (0.001 for NDF as it works better for it in the experiment) and train 10 epochs for each stage. The average test errors of the two different architectures are shown in Table~\ref{facetable}. Again, DHM supports PP to greatly reduce the computational cost and the performance only drops gracefully. This experiment validates again the strength of DHM over traditional NDF architecture in regression problems. Figure~\ref{result} shows some success and failure cases of this model. Compared with NDF, our DHM can significantly reduce the computational complexity after pruning  with even slightly better alignment accuracy.        

\section{Conclusion}
We proposed Deep Hierarchical Machine (DHM), a flexible framework for combining divide-and-conquer strategy and deep representation learning. Unlike recently proposed deep neural decision/regression forest, DHM can take advantage of the distribution of recommendation scores and a probabilistic pruning strategy is proposed to avoid unnecessary path evaluation. We also showed the feasibility of introducing sparse feature extraction process into DHM by using local binary convolution, which mimics traditional decision tree with pixel-difference feature and has potential for devices with limited computing resources.

{\small
\bibliographystyle{ieee}
\bibliography{reference}

\begin{thebibliography}{10}\itemsep=-1pt

\bibitem{2013978}
{\em Decision Forests for Computer Vision and Medical Image Analysis}.
\newblock Advances in Computer Vision and Pattern Recognition. Springer London,
  London, 2013.

\bibitem{8237378}
A.~Bulat and G.~Tzimiropoulos.
\newblock How far are we from solving the 2d amp; 3d face alignment problem?
  (and a dataset of 230,000 3d facial landmarks).
\newblock In {\em 2017 IEEE International Conference on Computer Vision
  (ICCV)}, pages 1021--1030, Oct 2017.

\bibitem{6248015}
X.~Cao, Y.~Wei, F.~Wen, and J.~Sun.
\newblock Face alignment by explicit shape regression.
\newblock In {\em 2012 IEEE Conference on Computer Vision and Pattern
  Recognition}, pages 2887--2894, June 2012.

\bibitem{109122}
D.~Chen, S.~Ren, Y.~Wei, X.~Cao, and J.~Sun.
\newblock Joint cascade face detection and alignment.
\newblock In D.~Fleet, T.~Pajdla, B.~Schiele, and T.~Tuytelaars, editors, {\em
  Computer Vision -- ECCV 2014}, pages 109--122, Cham, 2014. Springer
  International Publishing.

\bibitem{181214}
M.~I. Jordan and R.~A. Jacobs.
\newblock Hierarchical mixtures of experts and the em algorithm.
\newblock {\em Neural computation}, 6(2):181--214, 1994.

\bibitem{8099939}
F.~Juefei-Xu, V.~N. Boddeti, and M.~Savvides.
\newblock Local binary convolutional neural networks.
\newblock In {\em 2017 IEEE Conference on Computer Vision and Pattern
  Recognition (CVPR)}, pages 4284--4293, July 2017.

\bibitem{6909637}
V.~Kazemi and J.~Sullivan.
\newblock One millisecond face alignment with an ensemble of regression trees.
\newblock In {\em 2014 IEEE Conference on Computer Vision and Pattern
  Recognition}, pages 1867--1874, June 2014.

\bibitem{7410529}
P.~Kontschieder, M.~Fiterau, A.~Criminisi, and S.~R. Bulò.
\newblock Deep neural decision forests.
\newblock In {\em 2015 IEEE International Conference on Computer Vision
  (ICCV)}, pages 1467--1475, Dec 2015.

\bibitem{5540107}
C.~H. Lampert.
\newblock An efficient divide-and-conquer cascade for nonlinear object
  detection.
\newblock In {\em 2010 IEEE Computer Society Conference on Computer Vision and
  Pattern Recognition}, pages 1022--1029, June 2010.

\bibitem{7130626}
S.~Liao, A.~K. Jain, and S.~Z. Li.
\newblock A fast and accurate unconstrained face detector.
\newblock {\em IEEE Transactions on Pattern Analysis and Machine Intelligence},
  38(2):211--223, Feb 2016.

\bibitem{7298681}
B.~Liu, M.~Wang, H.~Foroosh, M.~Tappen, and M.~Penksy.
\newblock Sparse convolutional neural networks.
\newblock In {\em 2015 IEEE Conference on Computer Vision and Pattern
  Recognition (CVPR)}, pages 806--814, June 2015.

\bibitem{0269282}
S.~Masoudnia and R.~Ebrahimpour.
\newblock Mixture of experts: a literature survey.
\newblock {\em Artificial Intelligence Review}, 42(2):275--293, 2014.

\bibitem{111222}
J.~Park, S.~Li, W.~Wen, P.~T.~P. Tang, H.~Li, Y.~Chen, and P.~Dubey.
\newblock Faster cnns with direct sparse convolutions and guided pruning.
\newblock In {\em 2017 International Conference on Representation Learning
  (ICLR)}, November 2017.

\bibitem{6909614}
S.~Ren, X.~Cao, Y.~Wei, and J.~Sun.
\newblock Face alignment at 3000 fps via regressing local binary features.
\newblock In {\em 2014 IEEE Conference on Computer Vision and Pattern
  Recognition}, pages 1685--1692, June 2014.

\bibitem{7298672}
S.~Ren, X.~Cao, Y.~Wei, and J.~Sun.
\newblock Global refinement of random forest.
\newblock In {\em 2015 IEEE Conference on Computer Vision and Pattern
  Recognition (CVPR)}, pages 723--730, June 2015.

\bibitem{7384759}
S.~Ren, X.~Cao, Y.~Wei, and J.~Sun.
\newblock Face alignment via regressing local binary features.
\newblock {\em IEEE Transactions on Image Processing}, 25(3):1233--1245, March
  2016.

\bibitem{7780963}
A.~Roy and S.~Todorovic.
\newblock Monocular depth estimation using neural regression forest.
\newblock In {\em 2016 IEEE Conference on Computer Vision and Pattern
  Recognition (CVPR)}, pages 5506--5514, June 2016.

\bibitem{97458}
S.~R. Safavian and D.~Landgrebe.
\newblock A survey of decision tree classifier methodology.
\newblock {\em IEEE Transactions on Systems, Man, and Cybernetics},
  21(3):660--674, May 1991.

\bibitem{123456}
W.~Shen, Y.~Guo, Y.~Wang, K.~Zhao, B.~Wang, and A.~L. Yuille.
\newblock Deep regression forests for age estimation.
\newblock In {\em The IEEE Conference on Computer Vision and Pattern
  Recognition (CVPR)}, June 2018.

\bibitem{5995316}
J.~Shotton, A.~Fitzgibbon, M.~Cook, T.~Sharp, M.~Finocchio, R.~Moore,
  A.~Kipman, and A.~Blake.
\newblock Real-time human pose recognition in parts from single depth images.
\newblock In {\em CVPR 2011}, pages 1297--1304, June 2011.

\bibitem{2015}
F.~Solera, S.~Calderara, and R.~Cucchiara.
\newblock Learning to divide and conquer for online multi-target tracking.
\newblock In {\em The IEEE International Conference on Computer Vision (ICCV)},
  December 2015.

\bibitem{159232}
X.~Sun, Y.~Wei, S.~Liang, X.~Tang, and J.~Sun.
\newblock Cascaded hand pose regression.
\newblock In {\em The IEEE Conference on Computer Vision and Pattern
  Recognition (CVPR)}, June 2015.

\bibitem{8031057}
S.~Tulyakov, L.~A. Jeni, J.~F. Cohn, and N.~Sebe.
\newblock Viewpoint-consistent 3d face alignment.
\newblock {\em IEEE Transactions on Pattern Analysis and Machine Intelligence},
  40(9):2250--2264, Sept 2018.

\bibitem{321456}
P.~Viola and M.~Jones.
\newblock Rapid object detection using a boosted cascade of simple features.
\newblock In {\em Computer Vision and Pattern Recognition, 2001. CVPR 2001.
  Proceedings of the 2001 IEEE Computer Society Conference on}, volume~1, pages
  511--518. IEEE, 2001.

\bibitem{146155}
X.~Zhu, Z.~Lei, X.~Liu, H.~Shi, and S.~Z. Li.
\newblock Face alignment across large poses: A 3d solution.
\newblock In {\em Proceedings of the IEEE conference on computer vision and
  pattern recognition}, pages 146--155, 2016.

\end{thebibliography}
}

\end{document}